# Novel Artificial Human Optimization Field Algorithms – The Beginning

Satish Gajawada
Alumnus, Indian Institute of Technology Roorkee
Founder, Artificial Human Optimization Field
gajawadasatish@gmail.com

Hassan M.H. Mustafa
Faculty of Specified Education, Dept. of Educational Technology,
Banha University, Egypt
prof.dr.hassanmoustafa@gmail.com

**Abstract:** New Artificial Human Optimization (AHO) Field Algorithms can be created from scratch or by adding the concept of Artificial Humans into other existing Optimization Algorithms. Particle Swarm Optimization (PSO) has been very popular for solving complex optimization problems due to its simplicity. In this work, new Artificial Human Optimization Field Algorithms are created by modifying existing PSO algorithms with AHO Field Concepts. These Hybrid PSO Algorithms comes under PSO Field as well as AHO Field. There are Hybrid PSO research articles based on Human Behavior, Human Cognition and Human Thinking etc. But there are no Hybrid PSO articles which based on concepts like Human Disease, Human Kindness and Human Relaxation. This paper proposes new AHO Field algorithms based on these research gaps. Some existing Hybrid PSO algorithms are given a new name in this work so that it will be easy for future AHO researchers to find these novel Artificial Human Optimization Field Algorithms. A total of 6 Artificial Human Optimization Field algorithms titled "Human Safety Particle Swarm Optimization (HuSaPSO)", "Human Kindness Particle Swarm Optimization (HKPSO)", "Human Relaxation Particle Swarm Optimization (HRPSO)", "Multiple Strategy Human Particle Swarm Optimization (MSHPSO)", "Human Thinking Particle Swarm Optimization (HTPSO)" and "Human Disease Particle Swarm Optimization (HDPSO)" are tested by applying these novel algorithms on Ackley, Beale, Bohachevsky, Booth and Three-Hump Camel Benchmark Functions. Results obtained are compared with PSO algorithm.

**Keywords:** Artificial Humans, Artificial Human Optimization Field, Particle Swarm Optimization, Genetic Algorithms, Hybrid Algorithms, Global Optimization Techniques, Nature Inspired Computing, Bio-Inspired Computing, Artificial Intelligence, Machine Learning

**Highlights:** 1) World's First Hybrid PSO algorithm based on Human Kindness is proposed in this paper.
2) World's First Hybrid PSO algorithm based on Human Relaxation is proposed in this paper
3) World's First Hybrid PSO algorithm based on Human Disease is proposed in this paper
4) Made corrections to previous work under AHO Field in the Introduction Section of the paper
5) A Novel Section "Interesting Findings in Artificial Human Optimization Field" is present in this article
6) Some existing Hybrid PSO algorithms are given a new name in this paper

# 1 Introduction

A field is a particular branch of study. Artificial Human Optimization Field (AHO Field) is a latest field. Proposing a new algorithm is different from proposing a new field. Generally researchers propose new algorithms. But for the first time in research industry history, a young researcher proposed a new field through Transactions on Machine Learning and Artificial Intelligence journal paper. Artificial Human Optimization (AHO) is a very recent field which took its birth on December 2016. This work was published in Transactions on Machine Learning and Artificial Intelligence with title "Entrepreneur: Artificial Human Optimization". Hence this field is less than 2 years old. According to recent articles in AHO literature, there is scope for many PhD's and PostDoc's in Artificial Human Optimization Field (AHO Field). Also there exists an ocean of opportunities in Artificial Human Optimization Field. According to article "Entrepreneur: Artificial Human Optimization", the first article in AHO Field was proposed in 2012 and there exists less than 20 papers in AHO Field. This mistake was corrected and the first article in AHO Field was proposed in 2009 according to article "Artificial Human Optimization – An Introduction". Again there was a mistake. This mistake was corrected in article "Artificial Human Optimization – An Overview". The correction was that the first article in AHO Field was proposed in 2006. Again there was a mistake. Finally, according to this paper "Novel Artificial Human Optimization Field Algorithms – The Beginning", the first paper in AHO Field was proposed in 2003 with title "Society and Civilization: An Optimization Algorithm Based on the Simulation of Social Behavior". Also there exist more than 30 papers in AHO Field. According to a recent article in AHO Literature, there is scope for millions of research articles in AHO Field [1-12]. Papers [1-12] gives details about Artificial Human Optimization Field, its algorithms and its overview. Papers [13-17] shows Hybrid PSO algorithms which come under Artificial Human Optimization Field.

The rest of the article is organized as follows:
Section 2 shows Particle Swarm Optimization (PSO) algorithm. Section 3 to Section 8 shows "Human Safety Particle Swarm Optimization (HuSaPSO)", "Human Kindness Particle Swarm Optimization (HKPSO)", "Human Relaxation Particle Swarm Optimization (HRPSO)", "Multiple Strategy Human Particle Swarm Optimization (MSHPSO)", "Human Thinking Particle Swarm Optimization (HTPSO)" and "Human Disease Particle Swarm Optimization (HDPSO)" respectively. Interesting Findings in AHO Field are shown in Section 9. Section 10 gives results obtained. Finally, Conclusions are given in Section 11.

# 2 Particle Swarm Optimization

Particle Swarm Optimization (PSO) was proposed by Kennedy and Eberhart in 1995. PSO is based on Artificial Birds. It has been applied to solve complex optimization problems. Papers [18-24] shows you details related to PSO, its algorithms and its overview.

In PSO, first we initialize all particles as shown below. Two variables $pbest_i$ and gbest are maintained. $pbest_i$ is the best fitness value achieved by $i^{th}$ particle so far and gbest is the best fitness value achieved by all particles so far. Lines 4 to 11 in the below text helps in maintaining particle best and global best. Then the velocity is updated by rule shown in line no. 14. Line 15 updates position of i[th] particle. Line 19 increments the number of iterations and then the control goes back to line 4. This process of a particle moving towards its local best and also moving towards global best of particles is continued until termination criteria will be reached.

**Procedure:** Particle Swarm Optimization (PSO)

```
1) Initialize all particles
2) iterations = 0
3) do
4)      for each particle i do
5)          If ( f( xi ) < f( pbesti ) ) then
6)              pbesti = xi
7)          end if
8)          if ( f( pbesti ) < f( gbest ) ) then
9)              gbest = pbesti
10)         end if
11)     end for
12)     for each particle i do
13)         for each dimension d do
14)             vi,d = w*vi,d +
                      C1*Random(0,1)*(pbesti,d – xi,d)
                      + C2*Random(0,1)*(gbestd – xi,d)
15)             xi,d = xi,d + vi,d
17)         end for
18)     end for
19)     iterations = iterations + 1
20) while ( termination condition is false)
```

# 3 Human Safety Particle Swarm Optimization

In PSO particles move towards local best and global best. Almost all PSO algorithms are based on best location of particles. But there is another strategy which is moving towards the optimal by using worst location of particles. Some algorithms in PSO Field are based on this idea in which worst location of particles also helps in finding optimal solution.

The idea of using worst location of particles in the velocity updating equation was first introduced in [28]. A Novel PSO Algorithm was proposed in [25]. In this algorithm, a coefficient is calculated based on distance of particle to closest best and closest worst particles. This coefficient is used in updating velocity of particle. In [26], velocity is updated using both particles best and particles worst location. This work is extended in [27] where velocity of particle is updated using particles local worst, global worst, particles local best and global best of all particles. The velocity is updated in [11] where particles move towards the local best and global best in even iterations and move away from local worst and global worst in odd iterations.

According to our experience it can be observed that Humans not only learn from his/her own local best and other individuals global best but also learns from his/her own local worst and other individuals global worst. Hence in [28], a new PSO (NPSO) is proposed where optimal solution is found by moving away from local worst location of particle and global worst location of all particles. The algorithm (NPSO) proposed in [28] is based on Artificial Human Optimization Field Concepts because there are Humans who try to be on safe side by moving away from local worst and global worst. Hence NPSO in [28] is given a new name titled "Human Safety Particle Swarm Optimization (HuSaPSO)" in this current paper.

In line no. 14 in below procedure it can be seen that velocity update equation is based on moving away from local worst of particle and global worst of all particles. In NPSO work in [28], researchers haven't used inertia weight while updating velocity but in the below procedure, inertia weight is used. Human Safety Particle Swarm Optimization (HuSaPSO) is shown below:

**Procedure:** Human Safety Particle Swarm Optimization (HuSaPSO)

1) Initialize all particles
2) iterations = 0
3) **do**
4)     **for** each particle i **do**
5)         **If** ( $f(x_i) < f(pbest_i)$ ) **then**
6)             $pbest_i = x_i$
7)         **end if**
8)         **if** ( $f(pbest_i) < f(gbest)$ ) **then**
9)             $gbest = pbest_i$
10)         **end if**
11)     **end for**
12)     **for** each particle i **do**
13)         **for** each dimension d **do**
14)             $v_{i,d} = w*v_{i,d} + C_1*Random(0,1)*(x_{i,d} - pworst_{i,d}) + C_2*Random(0,1)*(x_{i,d} - gworst_d)$
15)             $x_{i,d} = x_{i,d} + v_{i,d}$
17)         **end for**
18)     **end for**
19)     iterations = iterations + 1
20) **while** ( termination condition is false)

## 4 Human Kindness Particle Swarm Optimization

There are no Hybrid PSO algorithms based on Human Kindness till date. Human Kindness is modeled by introducing $KindnessFactor_i$ for particle i. This factor is added in the position update equation in line number 15 of the below procedure. The more the KindnessFactor the faster is the movement of particle.

In this work, a random number between 0 and 1 is generated and assigned to KindnessFactor of particle. The Proposed Human Kindness Particle Swarm Optimization (HKPSO) is shown below:

**Procedure:** Human Kindness Particle Swarm Optimization (HKPSO)

1) Initialize all particles
2) iterations = 0
3) **do**
4)     **for** each particle i **do**
5)         **If** ( $f(x_i) < f(pbest_i)$ ) **then**
6)             $pbest_i = x_i$
7)         **end if**
8)         **if** ( $f(pbest_i) < f(gbest)$ ) **then**
9)             $gbest = pbest_i$
10)         **end if**
11)     **end for**
12)     **for** each particle i **do**
13)         **for** each dimension d **do**
14)             $v_{i,d} = w*v_{i,d} + C_1*Random(0,1)*(pbest_{i,d} - x_{i,d}) + C_2*Random(0,1)*(gbest_d - x_{i,d})$
15)             $x_{i,d} = x_{i,d} + KindnessFactor_i * v_{i,d}$
17)         **end for**
18)     **end for**
19)     iterations = iterations + 1
20) **while** ( termination condition is false)

## 5 Human Relaxation Particle Swarm Optimization

There are no Hybrid PSO algorithms based on Human Relaxation till date. All particles move in some direction in all iterations. There is nothing like relaxation for a particle. RelaxationProbability is introduced in this paper in an attempt to model Human Relaxation. A random number is generated in the line number 13 in the below procedure. If the random number generated is less than or equal to RelaxationProbability then the particle is said to be on relaxation state and this particle will skip velocity updating and position updating in this particular iteration. On the other hand, if the random number generated is greater than RelaxationProbability, then particle will undergo velocity and position updating just like in normal PSO. Proposed Human Relaxation Particle Swarm Optimization (HRPSO) is shown below:

**Procedure:** Human Relaxation Particle Swarm Optimization (HRPSO)

1) Initialize all particles
2) Initialize RelaxationProbability
2) iterations = 0

```
3) do
4)      for each particle i do
5)              If ( f( x_i ) < f( pbest_i ) ) then
6)                      pbest_i = x_i
7)              end if
8)              if ( f( pbest_i ) < f( gbest ) ) then
9)                      gbest = pbest_i
10)             end if
11)     end for
12)     for each particle i do
13)             if  Random(0,1) < = RelaxationProbability
14)                     continue   // continues to next particle
15)             end if
16)             for each dimension d do
17)                     v_{i,d} = w*v_{i,d} +
                                C_1*Random(0,1)*(pbest_{i,d} − x_{i,d})
                                + C_2*Random(0,1)*(gbest_d − x_{i,d})
18)                     x_{i,d} = x_{i,d} + v_{i,d}
19)             end for
20)     end for
21)     iterations = iterations + 1
22) while ( termination condition is false)
```

## 6 Multiple Strategy Human Particle Swarm Optimization

Hassan Satish Particle Swarm Optimization (HSPSO) proposed in [11] is given a new name titled "Multiple Strategy Human Particle Swarm Optimization (MSHPSO)" in this paper. MSHPSO is obtained by incorporation of Multiple Strategy Human Optimization (MSHO) concepts into Particle Swarm Optimization. In starting and even generations the Artificial Humans move towards the best fitness value. In odd generations Artificial Humans move away from the worst fitness value. In MSHPSO, local worst of particle and global worst of all particles are maintained in addition to local best of particle and global best of all particles. This is shown in lines 4 to 17. In lines 19 to 24 velocity is calculated by moving towards the local best of particle and global best of all particles. In lines 26 to 31 pseudo code for odd generations is shown in below text. In these odd generations particles move away from local worst of particle and also away from global worst of all particles. In line 33, number of iterations is incremented by one. Then control goes back to line number 4. This process of moving towards the best in one generation and moving away from the worst in next generation is continued until termination criteria has been reached. MSHPSO proposed in [11] is shown below:

**Procedure:** Multiple Strategy Human Particle Swarm Optimization (MSHPSO)

1) Initialize all particles
2) iterations = 0
3) **do**

```
4)      for each particle i do
5)           If ( f( x_i ) < f( pbest_i ) ) then
6)                pbest_i = x_i
7)           end if
8)           if ( f( pbest_i ) < f( gbest ) ) then
9)                gbest = pbest_i
10)          end if
11)          If ( f( x_i ) > f( pworst_i ) ) then
12)               pworst_i = x_i
13)          end if
14)          if ( f( pworst_i ) > f( gworst ) ) then
15)               gworst = pworst_i
16)          end if
17)     end for
18)     If ((iterations == 0) || (iterations%2==0)) then
                // for starting and even iterations
19)          for each particle i do
20)               for each dimension d do
21)                    v_{i,d} = w*v_{i,d} +
                              C_1*Random(0,1)*(pbest_{i,d} − x_{i,d})
                              +C_2*Random(0,1)*(gbest_d − x_{i,d})
22)                    x_{i,d} = x_{i,d} + v_{i,d}
23)               end for
24)          end for
25)     else // for odd iterations
26)          for each particle i do
27)               for each dimension d do
28)                    v_{i,d} = w*v_{i,d} +
                              C_1*Random(0,1)*( x_{i,d} - pworst_{i,d} )
                              + C_2*Random(0,1)*( x_{i,d} - gworst_d)
29)                    x_{i,d} = x_{i,d} + v_{i,d}
30)               end for
31)          end for
32)     end if
33)     iterations = iterations + 1
34) while ( termination condition is false)
```

## 7 Human Thinking Particle Swarm Optimization

In [12], the particles move towards best locations and away from worst locations in the same iteration/generation. The Concept used in [12] and [27] is same. The only difference is that a new name titled "Human Thinking Particle Swarm Optimization (HTPSO)" is given in [12] for the concept in [27].

Almost all Particle Swarm Optimization (PSO) algorithms are proposed such that the particles move towards best particles. But Human Thinking is such that they not only move towards best but also moves away from the worst. This concept was used to design algorithm titled "Multiple Strategy Human Optimization (MSHO)" in [4]. In MSHO, artificial Humans move towards the best in even generations and move away from the worst in odd generations. But in Human Thinking Particle Swarm Optimization, both strategies happen in the same generation and all generations follow the same strategy. That is moving towards the best and moving away from the worst strategies happen simultaneously in the same generation unlike MSHO designed in [4]. The HTPSO algorithm proposed in [12] and [27] is shown below:

**Procedure:** Human Thinking Particle Swarm Optimization (HTPSO)

1) Initialize all particles
2) iterations = 0
3) **do**
4)     **for** each particle i **do**
5)         **If** ( f( $x_i$ ) < f( $pbest_i$ ) ) **then**
6)             $pbest_i = x_i$
7)         **end if**
8)         **if** ( f( $pbest_i$ ) < f( gbest ) ) **then**
9)             gbest = $pbest_i$
10)         **end if**
11)         **If** ( f( $x_i$ ) > f( $pworst_i$ ) ) **then**
12)             $pworst_i = x_i$
13)         **end if**
14)         **if** ( f( $pworst_i$ ) > f( gworst ) ) **then**
15)             gworst = $pworst_i$
16)         **end if**
17)     **end for**
18)     **for** each particle i **do**
19)         **for** each dimension d **do**
20)             $v_{i,d} = w*v_{i,d} + Random(0,1)*(pbest_{i,d} - x_{i,d}) + Random(0,1)*(gbest_d - x_{i,d})$
21)             $v_{i,d} = v_{i,d} + Random(0,1)*( x_{i,d} - pworst_{i,d} ) + Random(0,1)*( x_{i,d} - gworst_d )$
22)             $x_{i,d} = x_{i,d} + v_{i,d}$
23)         **end for**
24)     **end for**
25)     iterations = iterations + 1
26) **while** (termination condition is false)

# 8 Human Disease Particle Swarm Optimization

In this section, an innovative Hybrid PSO algorithm titled "Human Disease Particle Swarm Optimization (HDPSO)" is proposed which is based on Bipolar Disorder Human Disease. People with Bipolar Disorder Human Disease experience changes in moods between depression and mania. This disease is also known as manic depression. The mood swings between highs of mania (very happy) and lows of depression (very sad) are significant and usually extreme. The mood of either mania or depression can exist for few days, few weeks or even few months.

In Human Disease Particle Swarm Optimization, the strategy for updating velocity is different in odd and even generations. Person affected with Bipolar Disorder Human Disease goes through Very happy (UP) and very sad (Down) phases. Very happy and Very sad phases of Bipolar Disorder Human Disease are imitated in proposed HDPSO algorithm by incorporating different updating strategies in Particle Swarm Optimization algorithm. If person is very happy then he moves towards global best and local best of particles. If the person is very sad then he moves away from global best and local best of particles. In line 15 in below procedure, the person moves towards global best and local best of particles. In line 22, the person moves away from global best and local best of particles. The proposed HDPSO algorithm is shown below:

**Procedure:** Human Disease Particle Swarm Optimization (HDPSO)

```
1) Initialize all particles
2) iterations = 0
3) do
4)     for each particle i do
5)         If ( f( xi ) < f( pbesti ) ) then
6)             pbesti = xi
7)         end if
8)         if ( f( pbesti ) < f( gbest ) ) then
9)             gbest = pbesti
10)        end if
11)    end for

12)    If ((iterations == 0) || (iterations%2==0)) then
                // for starting and even iterations
13)        for each particle i do
14)            for each dimension d do
15)                vi,d = w*vi,d +
                        C1*Random(0,1)*(pbesti,d − xi,d)
                        +C2*Random(0,1)*(gbestd − xi,d)
16)                xi,d = xi,d + vi,d
17)            end for
18)        end for
19)    else // for odd iterations
20)        for each particle i do
```

```
21)                    for each dimension d do
22)                        v_{i,d} = w*v_{i,d} +
                                C_1*Random(0,1)*( x_{i,d} - pbest_{i,d} )
                                + C_2*Random(0,1)*( x_{i,d} - gbest_d )
23)                        x_{i,d} = x_{i,d} + v_{i,d}
24)                    end for
25)                end for
26)        end if
27)        iterations = iterations + 1
28) while ( termination condition is false)
```

## 9 Interesting Findings in Artificial Human Optimization Field

Human Thinking Particle Swarm Optimization (HTPSO) was proposed in [12] by Satish Gajawada et al. in 2018. Velocity is updated in HTPSO such that particle moves towards its local best, global best of particles, its local worst and global worst of particles. But this idea of velocity update is already proposed in [27]. Hence from here on it should be noted that HTPSO algorithm in [12] was originally proposed in [27]. This mistake happened because researchers added concept of Artificial Humans into PSO algorithms but they have not included the word "Human" in naming the new algorithm or in the entire paper. Another reason is that there are common things between Artificial Birds and Artificial Humans. The idea in [12] and [27] belongs to this intersection. PSO researchers added this common behavior to PSO and considered it as an algorithm based on Artificial Birds. Artificial Human Optimization Field (AHO Field) is new and hence there are not many algorithms under this new field. It will be very difficult for an AHO researcher to find that his concept/algorithm already exists in the form of PSO variant.

The algorithm in [28] comes under Artificial Human Optimization Field. There are millions of articles on internet. It will be very difficult for Artificial Human Optimization (AHO) researcher to find the fact that there exists paper [28] which added human safety into PSO and hence this work in [28] comes under Artificial Human Optimization Field. Also the title of paper [28] is "New Particle Swarm Optimization Technique". Hence AHO researchers might think this is another algorithm inspired by birds. Hence from here on researchers should include word "Human" or some other word to know that it is a AHO concept algorithm.

## 10 Results

The results obtained after applying HuSaPSO, HKPSO, HRPSO, MSHPSO, HTPSO, HDPSO and PSO algorithms on various benchmark functions are shown in this section. The figures of benchmark functions are taken from [29].

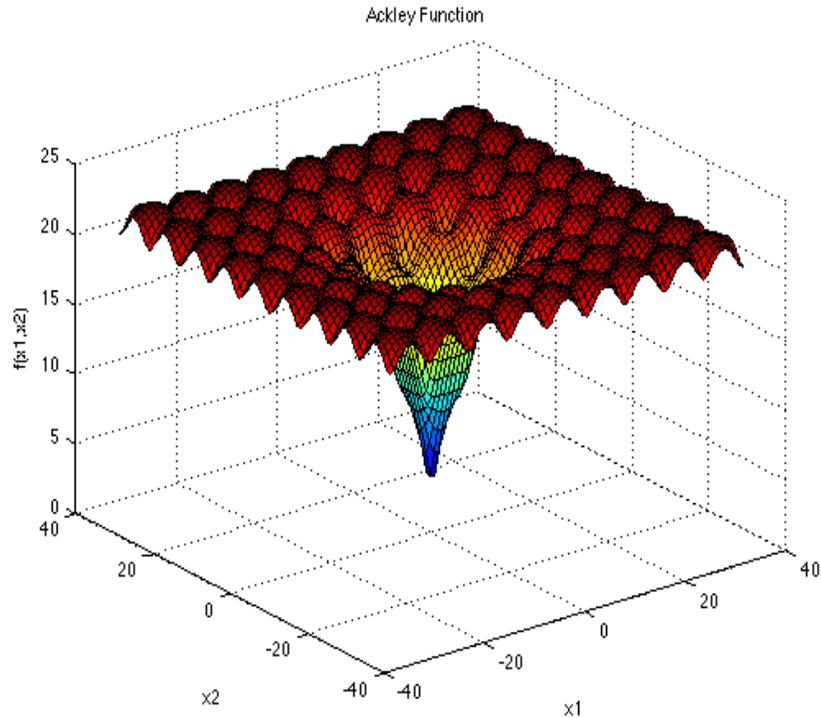

**Figure 1. Ackley Function**

```
C:\Users\qw\Desktop\Novel.AHO.Field.Algorithms\cdos.pso.HumanSafety>pso pso.run
begin time: Sun Nov 18 14:22:33 2018

0 run finished!
Best X :

 1.266626
 -1.588988
Optimal Value : 7.092738
end time: Sun Nov 18 14:22:33 2018
```

**Figure 2. Result given by Human Safety Particle Swarm Optimization (HuSaPSO) on Ackley Function**

```
C:\Users\qw\Desktop\Novel.AHO.Field.Algorithms\cdos.pso.HumanKindness>pso pso.ru
n
begin time: Sun Nov 18 18:51:26 2018

0 run finished!
Best X :

 -0.000000
 -0.000000
Optimal Value : 0.000000
end time: Sun Nov 18 18:51:26 2018
```

**Figure 3. Result given by Human Kindness Particle Swarm Optimization (HKPSO) on Ackley Function**

```
C:\Users\qw\Desktop\Novel.AHO.Field.Algorithms\cdos.pso.HumanRelaxation>pso pso.
run
begin time: Mon Nov 19 13:02:17 2018

0 run finished!
Best X :

 0.000000
 0.000000
Optimal Value : 0.000000
end time: Mon Nov 19 13:02:17 2018
```

**Figure 4. Result given by Human Relaxation Particle Swarm Optimization (HRPSO) on Ackley Function**

```
C:\Users\qw\Desktop\PSO.AHO\cdos.pso.modified>pso pso.run
begin time: Sun Jul 29 10:07:31 2018

0 run finished!
Best X :

 0.038762
 0.101817
Optimal Value : 0.597968
end time: Sun Jul 29 10:07:31 2018
```

**Figure 5. Result given by Multiple Strategy Human Particle Swarm Optimization (MSHPSO) on Ackley Function**

```
C:\Users\qw\Desktop\PSO.AHO\HTPSO\HTPSO.cdos.pso.modified>pso pso.run
begin time: Wed Jul 25 16:19:22 2018

0 run finished!
Best X :

 0.429100
 -0.591114
Optimal Value : 4.262748
end time: Wed Jul 25 16:19:22 2018
```

**Figure 6. Result given by Human Thinking Particle Swarm Optimization (HTPSO) on Ackley Function**

```
C:\Users\qw\Desktop\BipolarPSO\cdos.pso.bipolar.modified>pso pso.run
begin time: Tue Oct 30 16:19:27 2018

0 run finished!
Best X :

 -0.036502
 0.048344
Optimal Value : 0.266751
end time: Tue Oct 30 16:19:28 2018
```

**Figure 7. Result given by Human Disease Particle Swarm Optimization (HDPSO) on Ackley Function**

```
C:\Users\qw\Desktop\PSO.AHO\HTPSO\PSO.cdos>PSO PSO.RUN
begin time: Wed Jul 25 18:31:07 2018

0 run finished!
Best X :

 0.000000
-0.000000
Optimal Value : 0.000000
end time: Wed Jul 25 18:31:07 2018
```

**Figure 8. Result given by Particle Swarm Optimization (PSO) on Ackley Function**

From Figure 2 to Figure 8 it can be observed that HKPSO, HRPSO, PSO gave optimum solution and performed well on Ackley Function. But HuSaPSO, HTPSO, HDPSO, MSHPSO algorithms didn't perform well on Ackley Function.

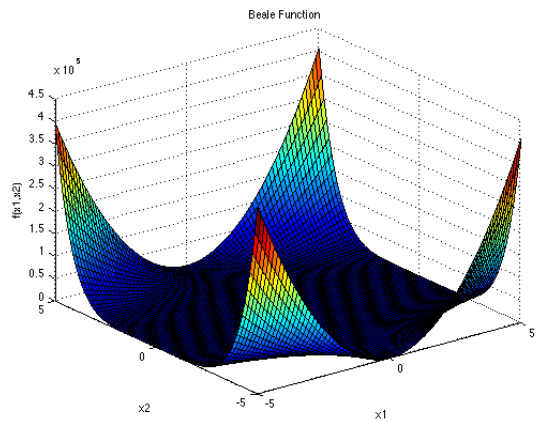

**Figure 9. Beale Function**

```
C:\Users\qw\Desktop\Novel.AHO.Field.Algorithms\cdos.pso.HumanSafety>pso pso.run
begin time: Sun Nov 18 14:31:14 2018

0 run finished!
Best X :

 2.891090
 0.319284
Optimal Value : 0.368583
end time: Sun Nov 18 14:31:15 2018
```

**Figure 10. Result given by Human Safety Particle Swarm Optimization (HuSaPSO) on Beale Function**

```
C:\Users\qw\Desktop\Novel.AHO.Field.Algorithms\cdos.pso.HumanKindness>pso pso.ru
n
begin time: Sun Nov 18 18:56:12 2018

0 run finished!
Best X :

 3.000000
 0.500000
Optimal Value : 0.000000
end time: Sun Nov 18 18:56:12 2018
```

**Figure 11. Result given by Huma Kindness Particle Swarm Optimization (HKPSO) on Beale Function**

```
C:\Users\qw\Desktop\Novel.AHO.Field.Algorithms\cdos.pso.HumanRelaxation>pso pso.
run
begin time: Mon Nov 19 13:05:07 2018

0 run finished!
Best X :

 3.000000
 0.500000
Optimal Value : 0.000000
end time: Mon Nov 19 13:05:07 2018
```

**Figure 12. Result given by Human Relaxation Particle Swarm Optimization (HRPSO) on Beale Function**

```
C:\Users\qw\Desktop\PSO.AHO\cdos.pso.modified>pso pso.run
begin time: Sun Jul 29 10:11:51 2018

0 run finished!
Best X :

 2.950198
 0.485876
Optimal Value : 0.000469
end time: Sun Jul 29 10:11:51 2018
```

**Figure 13. Result given by Multiple Strategy Human Particle Swarm Optimization (MSHPSO) on Beale Function**

```
C:\Users\qw\Desktop\PSO.AHO\HTPSO\HTPSO.cdos.pso.modified>PSO PSO.RUN
begin time: Wed Jul 25 17:52:43 2018

0 run finished!
Best X :

 2.729012
 0.332734
Optimal Value : 0.134325
end time: Wed Jul 25 17:52:43 2018
```

**Figure 14. Result given by Human Thinking Particle Swarm Optimization (HTPSO) on Beale Function**

```
C:\Users\qw\Desktop\BipolarPSO\cdos.pso.bipolar.modified>pso pso.run
begin time: Tue Oct 30 16:22:48 2018

0 run finished!
Best X :

 3.002366
 0.500054
Optimal Value : 0.000007
end time: Tue Oct 30 16:22:48 2018
```

**Figure 15. Result given by Human Disease Particle Swarm Optimization (HDPSO) on Beale Function**

```
C:\Users\qw\Desktop\PSO.AHO\HTPSO\PSO.cdos>PSO PSO.RUN
begin time: Wed Jul 25 18:34:03 2018

0 run finished!
Best X :

 3.000000
 0.500000
Optimal Value : 0.000000
end time: Wed Jul 25 18:34:03 2018
```

**Figure 16. Result given by Particle Swarm Optimization (PSO) on Beale Function**

From Figure 10 to Figure 16 it can be observed that HKPSO, HRPSO, MSHPSO, HDPSO and PSO gave optimal result and performed well on Beale Function. But HuSaPSO, HTPSO didn't perform well on Beale Function.

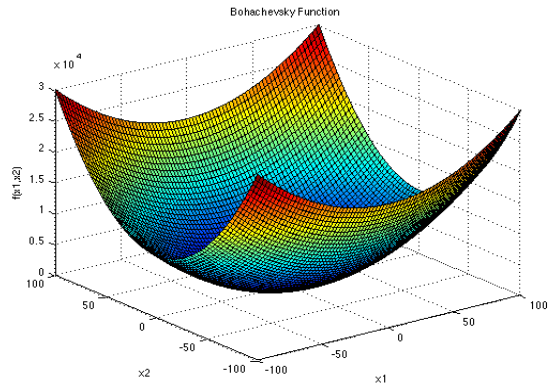

**Figure 17. Bohachevsky Function**

```
C:\Users\qw\Desktop\Novel.AHO.Field.Algorithms\cdos.pso.HumanSafety>pso pso.run
begin time: Sun Nov 18 14:36:50 2018

0 run finished!
Best X :

 4.222084
 -5.296627
Optimal Value : 75.115540
end time: Sun Nov 18 14:36:50 2018
```

**Figure 18. Result given by Human Safety Particle Swarm Optimization (HuSaPSO) on Bohachevsky Function**

```
C:\Users\qw\Desktop\Novel.AHO.Field.Algorithms\cdos.pso.HumanKindness>pso pso.ru
n
begin time: Sun Nov 18 18:58:57 2018

0 run finished!
Best X :

 0.000011
 -0.000004
Optimal Value : -0.000000
end time: Sun Nov 18 18:58:57 2018
```

**Figure 19. Result given by Human Kindness Particle Swarm Optimization (HKPSO) on Bohachevsky Function**

```
C:\Users\qw\Desktop\Novel.AHO.Field.Algorithms\cdos.pso.HumanRelaxation>pso pso.
run
begin time: Mon Nov 19 13:08:15 2018

0 run finished!
Best X :

 0.000000
 0.000000
Optimal Value : -0.000000
end time: Mon Nov 19 13:08:15 2018
```

**Figure 20. Result given by Human Relaxation Particle Swarm Optimization (HRPSO) on Bohachevsky Function**

```
C:\Users\qw\Desktop\PSO.AHO\cdos.pso.modified>pso pso.run
begin time: Sun Jul 29 10:16:59 2018

0 run finished!
Best X :

 0.620121
 0.056801
Optimal Value : 0.516828
end time: Sun Jul 29 10:16:59 2018
```

**Figure 21. Result given by Multiple Strategy Human Particle Swarm Optimization (MSHPSO) on Bohachevsky Function**

```
C:\Users\qw\Desktop\PSO.AHO\HTPSO\HTPSO.cdos.pso.modified>PSO PSO.RUN
begin time: Wed Jul 25 17:11:18 2018

0 run finished!
Best X :

 -1.322266
  1.193764
Optimal Value : 5.305778
end time: Wed Jul 25 17:11:18 2018
```

**Figure 22. Result given by Human Thinking Particle Swarm Optimization (HTPSO) on Bohachevsky Function**

```
C:\Users\qw\Desktop\BipolarPSO\cdos.pso.bipolar.modified>pso pso.run
begin time: Tue Oct 30 16:26:55 2018

0 run finished!
Best X :

  0.574630
 -0.062487
Optimal Value : 0.560673
end time: Tue Oct 30 16:26:55 2018
```

**Figure 23. Result given by Human Disease Particle Swarm Optimization (HDPSO) on Bohachevsky Function**

```
C:\Users\qw\Desktop\PSO.AHO\HTPSO\PSO.cdos>PSO PSO.RUN
begin time: Wed Jul 25 18:37:40 2018

0 run finished!
Best X :

 -0.000014
  0.000002
Optimal Value : -0.000000
end time: Wed Jul 25 18:37:40 2018
```

**Figure 24. Result given by Particle Swarm Optimization (PSO) on Bohachevsky Function**

From Figure 18 to Figure 24 it can be observed that HKPSO, HRPSO and PSO gave optimal result and performed well on Bohachevsky Function. But HuSaPSO, MSHPSO, HTPSO and HDPSO didn't perform well on Bohachevsky Function.

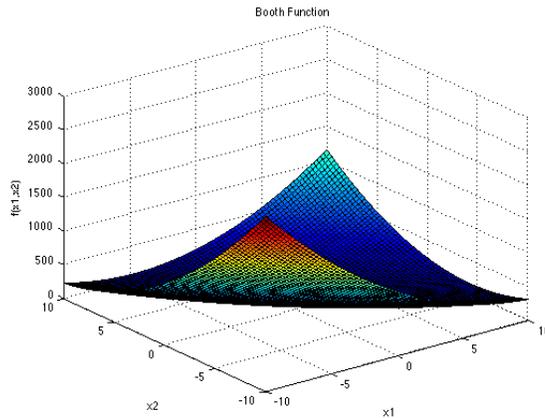

**Figure 25. Booth Function**

```
C:\Users\qw\Desktop\Novel.AHO.Field.Algorithms\cdos.pso.HumanSafety>pso pso.run
begin time: Sun Nov 18 14:41:00 2018

0 run finished!
Best X :

 0.845791
 3.062923
Optimal Value : 0.061072
end time: Sun Nov 18 14:41:00 2018
```

**Figure 26. Result given by Human Safety Particle Swarm Optimization (HuSaPSO) on Booth Function**

```
C:\Users\qw\Desktop\Novel.AHO.Field.Algorithms\cdos.pso.HumanKindness>pso pso.run
begin time: Sun Nov 18 19:01:17 2018

0 run finished!
Best X :

 1.000000
 3.000000
Optimal Value : 0.000000
end time: Sun Nov 18 19:01:17 2018
```

**Figure 27. Result given by Human Kindness Particle Swarm Optimization (HKPSO) on Booth Function**

```
C:\Users\qw\Desktop\Novel.AHO.Field.Algorithms\cdos.pso.HumanRelaxation>pso pso.run
begin time: Mon Nov 19 13:11:41 2018

0 run finished!
Best X :

 1.000000
 3.000000
Optimal Value : 0.000000
end time: Mon Nov 19 13:11:41 2018
```
**Figure 28. Result given by Human Relaxation Particle Swarm Optimization (HRPSO) on Booth Function**

```
C:\Users\qw\Desktop\PSO.AHO\cdos.pso.modified>pso pso.run
begin time: Sun Jul 29 10:18:55 2018

0 run finished!
Best X :

 0.994598
 3.024021
Optimal Value : 0.001993
end time: Sun Jul 29 10:18:55 2018
```

**Figure 29. Result given by Multiple Strategy Human Particle Swarm Optimization (MSHPSO) on Booth Function**

```
C:\Users\qw\Desktop\PSO.AHO\HTPSO\HTPSO.cdos.pso.modified>PSO PSO.RUN
begin time: Wed Jul 25 17:22:55 2018

0 run finished!
Best X :

 1.274603
 2.578953
Optimal Value : 0.338471
end time: Wed Jul 25 17:22:56 2018
```

**Figure 30. Result given by Human Thinking Particle Swarm Optimization (HTPSO) on Booth Function**

```
C:\Users\qw\Desktop\BipolarPSO\cdos.pso.bipolar.modified>pso pso.run
begin time: Tue Oct 30 16:30:10 2018

0 run finished!
Best X :

 0.986634
 3.009246
Optimal Value : 0.000332
end time: Tue Oct 30 16:30:10 2018
```

**Figure 31. Result given by Human Disease Particle Swarm Optimization (HDPSO) on Booth Function**

```
C:\Users\qw\Desktop\PSO.AHO\HTPSO\PSO.cdos>PSO PSO.RUN
begin time: Wed Jul 25 18:40:33 2018

0 run finished!
Best X :

 1.000000
 3.000000
Optimal Value : 0.000000
end time: Wed Jul 25 18:40:33 2018
```

**Figure 32. Result given by Particle Swarm Optimization (PSO) on Booth Function**

From Figure 26 to Figure 32 it can be observed that HuSaPSO gave result close to optimal solution and performed O.K. HKPSO, HRPSO, HDPSO, MSHPSO and PSO gave optimal result and performed well on Booth Function. But HTPSO didn't perform well on Booth Function.

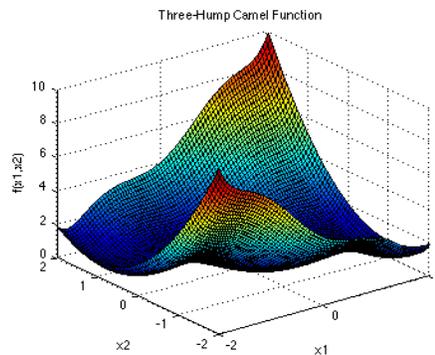

**Figure 33. Three-Hump Camel Function**

```
C:\Users\qw\Desktop\Novel.AHO.Field.Algorithms\cdos.pso.HumanSafety>pso pso.run
begin time: Sun Nov 18 14:45:18 2018

0 run finished!
Best X :

 0.211103
 -0.264831
Optimal Value : 0.101288
end time: Sun Nov 18 14:45:18 2018
```

**Figure 34. Result given by Human Safety Particle Swarm Optimization (HuSaPSO) on Three-Hump Camel Function**

```
C:\Users\qw\Desktop\Novel.AHO.Field.Algorithms\cdos.pso.HumanKindness>pso pso.ru
n
begin time: Sun Nov 18 19:04:07 2018

0 run finished!
Best X :

 0.000000
 0.000000
Optimal Value : 0.000000
end time: Sun Nov 18 19:04:07 2018
```

**Figure 35. Result given by Human Kindness Particle Swarm Optimization (HKPSO) on Three-Hump Camel Function**

```
C:\Users\qw\Desktop\Novel.AHO.Field.Algorithms\cdos.pso.HumanRelaxation>pso pso.
run
begin time: Mon Nov 19 13:14:30 2018

0 run finished!
Best X :

 0.000000
 0.000000
Optimal Value : 0.000000
end time: Mon Nov 19 13:14:30 2018
```

**Figure 36. Result given by Human Relaxation Particle Swarm Optimization (HRPSO) on Three-Hump Camel Function**

```
C:\Users\qw\Desktop\PSO.AHO\cdos.pso.modified>pso pso.run
begin time: Sun Jul 29 10:20:44 2018

0 run finished!
Best X :

 -0.008841
 -0.034073
Optimal Value : 0.001618
end time: Sun Jul 29 10:20:44 2018
```

**Figure 37. Result given by Multiple Strategy Human Particle Swarm Optimization (MSHPSO) on Three-Hump Camel Function**

```
C:\Users\qw\Desktop\PSO.AHO\HTPSO\HTPSO.cdos.pso.modified>PSO PSO.RUN
begin time: Wed Jul 25 17:36:55 2018

0 run finished!
Best X :

 -0.069427
 -0.093796
Optimal Value : 0.024926
end time: Wed Jul 25 17:36:55 2018
```

**Figure 38. Result given by Human Thinking Particle Swarm Optimization (HTPSO) on Three-Hump Camel Function**

```
C:\Users\qw\Desktop\BipolarPSO\cdos.pso.bipolar.modified>pso pso.run
begin time: Tue Oct 30 16:34:06 2018

0 run finished!
Best X :

 0.014122
 0.004653
Optimal Value : 0.000486
end time: Tue Oct 30 16:34:06 2018
```

**Figure 39. Result given by Human Disease Particle Swarm Optimization (HDPSO) on Three-Hump Camel Function**

**Figure 40. Result given by Particle Swarm Optimization (PSO) on Three-Hump Camel Function**

From Figure 34 to Figure 40 it can be observed that HKPSO, HRPSO, MSHPSO, HTPSO, HDPSO and PSO gave optimal result and performed well on Three-Hump Camel Function. But HuSaPSO didn't perform well on Three-Hump Camel Function.

| Benchmark Function / Algorithm | PSO | HuSaPSO | HKPSO | HRPSO | MSHPSO | HTPSO | HDPSO |
|---|---|---|---|---|---|---|---|
| Ackley | Green | Red | Green | Green | Red | Red | Red |
| Beale | Green | Red | Green | Green | Green | Red | Green |
| Bohachevsky | Green | Red | Green | Green | Red | Red | Red |
| Booth | Green | Blue | Green | Green | Green | Red | Green |
| Three-Hump Camel | Green | Red | Green | Green | Green | Green | Green |

**Figure 41. Overall Result**

In Figure 41 first row shows PSO algorithms and first column shows benchmark functions. Green represents "Performed Well". Red represents "Didn't Performed Well". Blue represents "Performed O.K." or "Performed Between Well and Not Well".

From above figure it is clear that HKPSO, HRPSO and PSO performed Well for all benchmark functions whereas HuSaPSO didn't perform well even on single benchmark function. MSHPSO and HSPSO performed well on three benchmark functions. HTPSO performed well on only single benchmark function.

# 11 Conclusions

Hybrid PSO algorithms inspired by Human Kindness (HKPSO), Bipolar Disorder Human Disease (HDPSO) and Human Relaxation (HRPSO) are proposed in this novel work. Two previous Hybrid PSO algorithms are given a new name titled "Human Safety Particle Swarm Optimization (HuSaPSO)" and "Multiple Strategy Human Particle Swarm Optimization (MSHPSO)" in this research paper. A total of 7 algorithms are applied on set of 5 benchmark functions and results obtained are shown in this work. It can be concluded that just because some optimization algorithm is inspired by Humans doesn't mean it will perform better than other optimization algorithms like optimization algorithms inspired by other beings like Birds (PSO). It can be observed from this work that some AHO algorithms performed as good as PSO where as some other AHO algorithms didn't perform as good as PSO. This is just the beginning of research in Artificial Human Optimization Field (AHO Field).

## Satish Gajawada's Profile

**Satish Gajawada** completed his studies from world class institute "Indian Institute of Technology Roorkee (IIT Roorkee)". He is called as **"Father of Artificial Human Optimization Field"** by few experts for his valuable contribution to new field titled "Artificial Human Optimization". He received a **SALUTE and APPRECIATION from IEEE chair Dr. Eng. Sattar B. Sadkhan** for his numerous achievements within the field of science. Invited by WDD 2019 China to deliver a speech on "Artificial Human Optimization Field". Invited by DISP 2019 United Kingdom to deliver a Keynote talk titled "Artificial Human Optimization – An Overview". Published 25 research articles by the age of 30 years. Articles of Satish Gajawada are **featured in AI Today Science Magazine in 2018**. Below are the links:

https://www.aitoday.xyz/artificial-human-optimization-an-introduction/
https://www.aitoday.xyz/an-ocean-of-opportunities-in-artificial-human-optimization-field/
https://www.aitoday.xyz/collection-of-abstracts-in-artificial-human-optimization-field/
https://www.aitoday.xyz/hide-human-inspired-differential-evolution-an-algorithm-under-artificial-human-optimization-field/
https://www.aitoday.xyz/25-reviews-on-artificial-human-optimization-field-for-the-first-time-in-research-industry/

His work was Published in **AI Today Science Magazine** on **July, 2018.** Below is the link:

https://www.aitoday.xyz/artificial-human-optimization-an-overview/

All India Rank **917 in IIT-JEE 2007**. An awardee of MHRD scholarship by scoring **98.47 percentile in GATE 2011**. He is the author of "Artificial Human Optimization – An Introduction" and many other articles. In December 2016, he proposed a new field titled "Artificial Human Optimization" which comes under Artificial Intelligence. This work was published in "Transactions on Machine Learning and Artificial Intelligence". He got reviews like "very interesting", "very impressive" etc. for his work under Artificial Human Optimization Field.